\documentclass{article}
\usepackage{graphicx} 
\usepackage{amsmath}
\usepackage{amssymb}
\usepackage{xcolor}
\usepackage{tikz}

\usepackage[margin=1.2in]{geometry}

\newtheorem{theorem}{Theorem}[section]
\newtheorem{lemma}[theorem]{Lemma}
\newtheorem{proposition}[theorem]{Proposition}
\newtheorem{corollary}[theorem]{Corollary}

\newtheorem{remark}{Remark}[section]

\newcommand{\R}{\mathbb{R}}
\newcommand{\E}{\mathbb{E}}
\newcommand{\PP}{\mathbb{P}}

\newcommand{\Spos}[1]{\mathbb{S}^{#1}_{+}}
\newcommand{\Sp}[1]{\mathbb{S}^{#1}}

\title{
Randomly initialized autoencoders: \\
fixed points and edge-of-chaos
}

\begin{document}

\author{L. Berlyand, R. Sarapin, Y. Shmalo, V. Slavin, S. Sodin}

\maketitle

\begin{abstract}

In this paper we study autoencoders, a special class of deep neural nets (DNNs) whose performance can be characterized via their fixed points. This perspective naturally raises questions of  existence, stability, and basins of attraction of these fixed points. These questions are addressed via the contractive properties of autoencoders, and are closely related to the notion of edge-of-chaos.

Edge-of-chaos (EoC) is an important notion in the theory of DNNs. It describes the critical regime separating ordered and chaotic signal propagation through a randomly initialized network. Initialization at or near this critical regime offers several theoretical and practical advantages, including stability of the network w.r.t.\ perturbations of the input. EoC was previously introduced for broad classes of neural networks using mean-field averaging methods.
In this paper we modify the notion of EoC for the study of autoencoders. Specifically, we introduce local and global EoC for autoencoders that controls local (small) and global (arbitrary) perturbations of the input respectively.

The study of stability of autoencoders falls within the scope of  nonlinear problems in Random Matrix Theory (RMT).
Our analysis of local EoC is based on spectral techniques of RMT, whereas global EoC is studied by employing Sudakov–Fernique inequality for Gaussian processes.

\end{abstract}

\section{Introduction}

\subsection{Autoencoder DNNs: what and why?}

Autoencoders (AEs) constitute an important class of deep neural networks (DNNs) designed to learn efficient representations of data (e.g., image) through compression and reconstruction.
We first recall the definition 
of a feedforward fully-connected DNN. 
For each $k=1,2,\ldots, L$, the $k$-th layer function is
\begin{equation}\label{def_Phi_DNN}
\Phi^{(k)}:\ \mathbb{R}^{n_k}\to\mathbb{R}^{n_{k+1}},\quad  \Phi^{(k)}(s)=\lambda(W^{(k)}s+b^{(k)}),
\end{equation}
where  $W^{(k)}$ is an ${n_{k+1} \times {n_k}}$  matrix, $b^{(k)} \in \mathbb R^{n_{k+1}}$ is a vector, and $\lambda$ is a nonlinear  function applied coordinatewise. In other words, for each $k$, $\Phi^{(k)}$ is a composition of an affine map $s\mapsto W^{(k)}s+b^{(k)}$  and a nonlinear function $\lambda\colon \mathbb{R}\to\mathbb{R}$ applied to each coordinate of $W^{(k)}s+b^{(k)}$.  The matrix $W^{(k)}$ is called a \emph{weight matrix}, the vector $b^{(k)}$ is called a \emph{bias vector}, and the function $\lambda$ is called an \textit{activation function}.

An $L$-layer DNN function is defined as a composition:
\begin{equation}\label{eq:DNN}
{\bf X}(s)=(\Phi^{(L)}\circ\Phi^{(L-1)}\circ\ldots\circ \Phi^{(2)}\circ\Phi^{(1)})(s).
\end{equation}
The vectors $s\in\mathbb{R}^{n_1}$ and $\bar{s}={\bf X}(s)\in\mathbb{R}^{n_{L+1}}$  are called \textit{input} and \textit{output vectors} correspondingly.
 
Autoencoders (AEs) are a special class of DNNs when the input and output vectors  have the same dimension $n_1=n_{L+1}=N$. An autoencoder 
consists of two DNN functions: an encoder 
${\bf X}_E$, and a decoder ${\bf X}_D$, i.e. 
\begin{equation}
{\bf X}_{AE} = {\bf X}_D\circ{\bf X}_E.
\label{AE}
\end{equation}
\noindent The encoder ${\bf X}_E$ maps the input $\boldsymbol{s}$ into a lower-dimensional latent representation $\boldsymbol{z} \in \mathbb{R}^{n}$, with $n \ll N$,
while the decoder ${\bf X}_D$ reconstructs the original input from this latent space:
\begin{equation}
{\bf X}_E(\boldsymbol{s}) = \boldsymbol{z},\quad{\bf X}_D(\boldsymbol{z}) = \boldsymbol{\bar s}.
\label{AE_Decoder}
\end{equation}

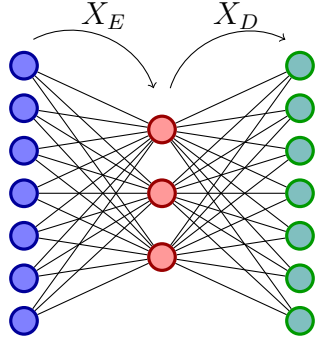
\begin{figure}[ht]
	\centering
	
	\begin{tikzpicture}[x=4pt,y=4pt]
		\node at (0,28) {};
		\node (v1) at (0,26) {};
		\node at (7.5,28.7) {\large $X_E$};
		\node at (20,28.7) {\large $X_D$};
		\node (u1) at (12.5, 21) {};
		\node (w1) at (13.5, 21) {};
		\node (t1) at (25.5, 26) {};
		\node (x1) at (18.5, 22) {};
		\node (y1) at (26.5, 18) {};
		\tikzstyle{every node}=[draw,shape=circle,minimum size=10pt, very thick];
		\foreach \x in {0,4,8,12,16,20,24} 
		{
			\node (a\x)[color=blue!60!black, fill=blue!50] at (0,\x) {};
		}
		
		
		\foreach \x in {6,12,18} 
		{
			\node (b\x)[color=red!60!black, fill=red!40] at (13,\x) {};
			\foreach \y in {0,4,8,12,16,20,24} 
			{
				\draw[color=black,thin] (a\y) -- (b\x);
			}
		}
		
		\foreach \x in {0,4,8,12,16,20,24} 
		{
			\node (c\x)[color=green!60!black, fill=teal!50] at (26,\x) {};
			\foreach \y in {6,12,18} 
			{
				\draw[color=black,thin] (b\y) -- (c\x);
			}
		}
		\draw[->]  (v1) to [bend left=50] (u1);
		\draw[->]  (w1) to [bend left=50] (t1);
	\end{tikzpicture}
	\caption{\textit{A simple AE with $N=7$, $n=3$. The blue and green nodes are the components of the input and output vectors, the red nodes form the latent layer.}}
	\label{fig1}
\end{figure}

A simple AE with a one-layer encoder and decoder is represented in Fig.~\ref{fig1}.

Autoencoders (AEs) have become a versatile tool in modern data analysis due to their ability to
learn compact and informative representations of high dimensional data in an unsupervised manner.
One application of AEs is nonlinear dimensionality reduction, where AEs generalize classical
techniques such as principal component analysis, see, e.g., \cite{Ge-Co:2018}. 
In this case AE function $X_{AE}$ (see \eqref{AE} - \eqref{AE_Decoder}) performs sequential mapping of input vector (image) $\boldsymbol{s}$ to latent layer vector $\boldsymbol{z}$ and then mapping of $\boldsymbol{z}$ back to $\boldsymbol{s}$, i.e., $X_{AE}$  acts as identity function.  Due to the fact that the latent layer is smaller or even much smaller than input and output layers, AE function cannot provide identity map on the entire $\R^N$. Our goal is to provide a map which is small deviation from identity map on a low-dimensional manifold. If AE function $X_{AE}$ provides contraction on a subset $S$ of the manifold, then according to the Banach fixed point  theorem there exist unique fixed point in $S$:
\begin{equation}\label{AE_FP}
{\bf X}_{AE}(s) = s
\end{equation}
with basin of attraction $S$. These fixed points correspond to images which are compressed and decompressed precisely, and the rest of input vectors from $S$ form a class which can be restored approximately. 
Hence, fixed points can be used as unique identifier (label) for all the input vectors lying in the basin of attraction.
For this reason, the criteria of  $X_{AE}$ contraction are important for the AE design and are studied in this work.

Weight matrices $W^{(k)}$ and bias vectors $b^{(k)}$  in \eqref{def_Phi_DNN} are determined by the choice of their entries. This process is called {\it initialization}.  
However, for a given choice of $W^{(k)}$ and $b^{(k)}$, AE may not have fixed points,  and those appear after iterative adjustment of weights, see \cite{Be-Sla:2026}.
This procedure is called {\it training} and it amounts to minimization of the so-called {\it loss function}, e.g., ~\cite{Ber-Jab:2023,Goodfellow:2016}.

In the problem of image encoding/decoding the loss function can be chosen as:
\begin{equation}
{\cal L}(\{\alpha\}) = \sum_{s\in T}|{\bf X}_{AE}(\boldsymbol{s}, \{\alpha\}) - \boldsymbol{s}|,
\label{AE_Loss}
\end{equation}
where $T\subset S$ is a so-called training set.
\noindent Here $\{\alpha\}$ denotes the set of the entries of all weight matrices  and bias vectors.

The efficiency, e.g. speed and accuracy, of training critically depends on the initialization \cite{Glo-Ben:2010}, and it was shown (e.g. \cite{Schmid:2015,Sca-Bru:2024}) that the best training results can be achieved when the matrices are initialized\textit{ randomly}. 
Our main focus is how the choice of random initialization determines the existence, stability, and basins of attraction of the fixed points of AEs. In this paper we consider AEs initialized with centered Gaussian weights and study the dependence of the behavior on the variance of the weights.

Particular attention in the work is paid to the study of edge-of-chaos (EoC) for general feed-forward DNNs (see,  e.g.,\ \cite{Poole:2016, Pe-Ga:2018, Sa-Ta:2023, Ha:2018, De-Ta:2026}). This is a critical regime of parameter initialization which corresponds to the a transition between two DNN behaviors: stability and chaos.
When the initialization is chosen close to EoC, small perturbations in the input data neither vanish nor explode while the input propagates through the DNN, which leads to the most efficient transfer of information through the layers. In the case of Gaussian initialization with the weights distributed as $N(0,\sigma^2/n)$, EoC is the critical value $\sigma^2_{crit}$ of the variance parameter $\sigma^2$ such that for $\sigma^2<\sigma^2_{crit}$ perturbations of the input vanish and for $\sigma^2>\sigma^2_{crit}$ perturbations explode while propagating through the layers.

The focus of our paper is on autoencoder DNNs. Therefore, we introduce the notion of  EoC specifically for autoencoders that allow one to study existence, stability, and basins of attraction of fixed points. Moreover,  stability of fixed points can be studied for small perturbations (local stability) or arbitrary perturbations (global stability). For this reason we introduce local and global EoC for autoencoders in discussions after Theorem~\ref{thm:main_loc} and Corollary~\ref{cor:basin} respectively. 
Note that in the study of fixed points in neuroscience science literature for firing networks (see, e.g.,  \cite{Wa-To:2013, Zh-Ro:2023, Ri:2024, Sch-Os-Ab:2013}), it is sufficient to consider local stability. For autoencoders, uniform convergence w.r.t.\ inputs is important for the global stability.

In the local stability analysis, \textit{Random Matrix Theory }(RMT) techniques can be successfully applied to study input-output Jacobian 
of the autoencoder function. However, for the global stability the uniformity in input $s$ is required, which is essentially harder to capture using standard RMT approaches. Instead, we develop novel techniques based on the theory of Gaussian processes and \textit{Sudakov-Fernique inequality }(Theorem~\ref{thm:SF}, see \cite{Sud:1971,Sud:1976,Fern:1997}).

\subsection{Notation and main results}

The principal object of our study is an $L$-layer autoencoder ${\bf X}^{(N)}_{L,\sigma^2}(s)$ (\ref{eq:DNN}) with
\[
\Phi_k (s)  = \phi_\alpha(W^{(k)}s + b^{(k)})~,  \]
where
$W^{(k)}$ is an ${n_{k+1} \times {n_k}}$ random matrix with $N\left(0,\dfrac{\sigma^2}{n_k}\right)$-distributed entries,  $b^{(k)}\in\mathbb{R}^{n_{k+1}}$  is a random vector with  $N(0, \tilde \sigma^2)$, all the random variables are jointly independent,  input and output dimensions are  equal: $n_1=n_{L+1}=N$, and  the activation function $\phi_\alpha$ ($0 \leq \alpha < 1$) is given by
\begin{equation}
\phi_\alpha(x)=
\begin{cases}
x, & x\ge 0;\\
\alpha x & x<0,
\end{cases}
\end{equation} 
(known as ReLU for $\alpha = 0$ and LeakyReLU for $0 < \alpha < 1$).  Let 
\begin{equation}\label{IO_def}
{\bf J}_{L,\sigma^2}^{(N)}(s)=\dfrac{\partial {\bf X}^{(N)}_{L,\sigma^2}(s)}{\partial s}
\end{equation}
be the \textit{input-output Jacobian} of the neural network. Our first result describes the typical behavior of $\|{\bf J}_{L,\sigma^2}^{(N)}(s)\|$ for a fixed vector $s$.

\begin{theorem}\label{thm:main_loc} Assume $\alpha = 0$ (ReLU activation). 
Assume that for each $k=2,3,\ldots,L$ there exist $\lim\limits_{N\to\infty} \dfrac{n_k}{N}=c_k\in (c;C)$ for some $C>c>0$.
Then, for any $s \in \mathbb R^N$, we have almost surely
\begin{equation}
\lim_{L \to \infty} \lim_{N \to \infty} \frac{1}{L} \log \| {\bf J}_{L,\sigma^2}^{(N)}(s) \| = \frac12 \log (\sigma^2 / 2)~.
\end{equation}

\end{theorem}

To recast this result in the form reminiscent of the DNN terminology, we can define the {\em local edge-of-chaos of autoencoder}  as the critical value $\sigma^2_{L, \text{crit}, \text{loc}}$ for which almost surely
\begin{equation}\label{conv_log_norm_0}
\lim_{N \to \infty} \frac{1}{L} \log \| {\bf J}_{L,\sigma^2}^{(N)}(s) \| = 0~.
\end{equation}
Then Theorem~\ref{thm:main_loc} can be equivalently stated as 
\begin{equation}\label{sigma_loc_to_2}
\lim_{L \to \infty} \sigma^2_{L, \text{crit}, \text{loc}} = 2~.
\end{equation}

The notion of edge-of-chaos was introduced and studied for a wide class of DNNs, see e.g.,\ \cite{Poole:2016, Pe-Ga:2018, Sa-Ta:2023, Ha:2018, De-Ta:2026}, and is defined as the critical  value $\sigma^2_{\mathrm{crit}}$ of the variance $\sigma^2$ in the initialization $N(0,\dfrac{\sigma^2}{n_k})$ for which
\begin{equation}\label{converg:def_loc_av_eoc}
\mathbb{E}\Bigl\{\dfrac{1}{N}\mathrm{Tr}\, ({\bf J}^{(k)}(s))^T{\bf J}^{(k)}(s)\Bigr\} \to 1
\quad
\text{as $N\to\infty$} ,
\end{equation}
where ${\bf J}^{(k)}(s)$ is the input-output Jacobian of the $k$-th layer function $\Phi^{(k)}$ of the DNN. Note that $\sigma^2_{\mathrm{crit}}$ is independent of $k$ because of the assumption that the diagonal components of Jacobian are independent of  $k$, see e.g., \cite[Section 2.2]{Pe-Ga:2018} and \cite[Section 3]{Poole:2016}.

The edge-of-chaos introduced in e.g.,\ \cite{Poole:2016, Pe-Ga:2018, Sa-Ta:2023, Ha:2018, De-Ta:2026} shows whether small (local) perturbations of the input data are dying out or exploding when propagating through multilayer DNN, while our notion of local EoC shows if a multilayer DNN has a local contraction property (stability w.r.t.\ small perturbations). In contrast, the global EoC introduced below shows if a multilayer DNN has a contraction property on the entire set of data (stability w.r.t.\ arbitrary perturbations).

Note that  it was proved in \cite[Subsection 3.4]{Pe-Ga:2018} that for $\sigma^2=\sigma^2_{\mathrm{crit}}$ convergence~\eqref{converg:def_loc_av_eoc} also holds with ${\bf J}_{L,\sigma^2}^{(N)}$ in place of ${\bf J}^{(k)}$, i.e.\ edge-of-chaos $\sigma^2_{\mathrm{crit}}$ can be defined in terms of the input-output Jacobian of the entire DNN rather than single layer. Then the standard inequality between Hilbert-Schmidt and operator norms implies 
\begin{equation}
\sigma^2_{L,\mathrm{crit,\,loc}}\le \sigma^2_{\mathrm{crit}}.
\end{equation}
Moreover, since for ReLU activation $\sigma^2_{\mathrm{crit}}=2$ (it was proved for DNNs with layers of the same width, see \cite[Table 1]{Pe-Ga:2018}), our result~\eqref{sigma_loc_to_2} yields
\begin{equation}\label{sigma_loc_to_2}
\lim_{L \to \infty} \sigma^2_{L, \text{crit}, \text{loc}} = \sigma^2_{\mathrm{crit}}~
\end{equation}
for ReLU activation, that is for large number of layers these two values of EoC coincide.
\medskip

An important issue in applications of autoencoders is to prove that~\eqref{conv_log_norm_0} holds uniformly in $s$, which will imply that for $\sigma^2<\sigma^2_{\mathrm{crit}}$ autoencoder ${\bf X}^{(N)}_{L,\sigma^2}$ is a contraction not only on a neighborhood of a given input $s$, but also on the entire $\mathbb{R}^N$.
\medskip

The following second result  provides a bound which is uniform in the initial vector $s$. In this theorem we allow arbitrary $0 \leq \alpha < 1$. On the other hand,
we consider the particular case of no biases, i.e.\ $\tilde \sigma^2 = 0$. In this case, it is sufficient to control $\|{\bf X}(s)\|$ rather than $\|{\bf J}(s)\|$ when studying global stability of a fixed point, see Corollary~\ref{cor:basin} below.

\begin{theorem}\label{thm:main_glob} 
Assume $0 \leq \alpha < 1$,  $\tilde \sigma^2 = 0$ (no bias), and ${\bf X}^{(N)}_{L,\sigma^2}(s)$ has layer widths $cN\le n_k\le CN$ for some $C>c>0$. Denote $c_k=\dfrac{n_k}{N}\in[c;C]$ and
\begin{equation}\label{sig_+_1}
\beta_\alpha^{(L)}
=
\begin{cases}
\sqrt{2}L\left(1+\sum\limits_{k=1}^{L-1}\sqrt{1+\dfrac{c_{k+1}}{c_{k}}}+\sqrt{\dfrac{1}{c_L}}\right)^{-1},
\quad\text{if $\alpha=0$;}\\
\noalign{\vskip9pt}
L\left(1+\sum\limits_{k=1}^{L-1}\sqrt{1+\dfrac{c_{k+1}}{c_{k}}\cdot\dfrac{1+\alpha^2}{2}}+\sqrt{\dfrac{1}{c_L}}\cdot\sqrt{\dfrac{1+\alpha^2}{2}}\right)^{-1},
\quad\text{if $0<\alpha<1$.}
\end{cases}
\end{equation} 
Then for any $\varepsilon>0$ there exists $\widetilde{c}=\widetilde{c}(\varepsilon,L,c,C)>0$ such that with probability $\geq 1-\exp\{-\widetilde{c}NL\}$:
\begin{equation}\label{prob_norm<1}
\sup_{\substack{s\in\Omega_\alpha\\ \|s\|=1}} \|{\bf X}_{L,\sigma^2}^{(N)}(s)\|\leq\left(\dfrac{\sigma}{\beta_\alpha^{(L)}}+\varepsilon\right)^L,
\end{equation}
where
\begin{equation}\label{def_Omega_alpha}
\Omega_\alpha=
\begin{cases}
\mathbb{R}^N_+,\quad\text{if $\alpha=0$;}\\
\noalign{\vskip4pt}
\mathbb{R}^N,\quad\text{if $0<\alpha<1$,}
\end{cases}
\qquad
\mathbb{R}^N_+=\{(s_1,\ldots,s_N)\in\mathbb{R}^N\colon s_1,\ldots,s_N\ge 0\}.
\end{equation}
\end{theorem}

The following two corollaries describe a special case of Theorem~\ref{thm:main_glob} which captures asymptotic behavior of autoencoders in the large-depth limit.
Specifically, it provides a simple upper bound for the variance in the distribution of random initialization that insures that the autoencoder has unique stable fixed point.

\begin{corollary}
\label{cor:aut_L_to_inf} 
Assume $0 \leq \alpha < 1$,  $\tilde \sigma^2 = 0$ (no bias), and ${\bf X}^{(N)}_{L,\sigma^2}$ has layer widths $n_1=n_2=\ldots=n_{l-1}=n_{l+1}=\ldots=n_{L+1}=N$, $cN\le n_l<N$ for some $l\in[2;L]$ and fixed $c\in (0;1)$.
Denote
\begin{equation}\label{def_beta_alpha}
\beta_\alpha=
\begin{cases}
1,\quad\text{if $\alpha=0$;}\\
\noalign{\vskip7pt}
\sqrt{\dfrac{2}{3+\alpha^2}},\quad\text{if $0<\alpha<1$.}
\end{cases}
\end{equation}
 If 
$
\sigma < \beta_\alpha
$,
then with probability $\geq 1 - \widetilde{C} \exp (- \widetilde{c} N)$ (where $\widetilde{C}, \widetilde{c} > 0$ depend on $\sigma^2, c$) we have:
\begin{equation}
 \lim_{L \to \infty}  \sup_{\stackrel{s\in\Omega_\alpha}{\scriptscriptstyle\|s\|=1}} \| {\bf X}_{L,\sigma^2}^{(N)}(s) \| = 0~,\end{equation}
where $\Omega_\alpha$ defined as in~\eqref{def_Omega_alpha}.
\end{corollary}

Due to Corollary~\ref{cor:aut_L_to_inf}, if $\sigma<\beta_\alpha$ then for large enough $L\geq\widetilde{L}$ with probability $\geq 1 - \widetilde{C} \exp (- \widetilde{c} N)$ we have
\begin{equation}\label{sup_norm<K}
K:=\sup_{\stackrel{\scriptstyle s\in\Omega_\alpha}{\scriptstyle\|s\|=1}}\|{\bf X}_{L,\sigma^2}^{(N)}(s)\|<1.
\end{equation}
Consider an iterative process
\begin{equation}\label{iter_proc}
s_0\in\mathbb{R}^N,\quad
s_m={\bf X}_{L,\sigma^2}^{(N)}(s_{m-1}).
\end{equation}
Notice that $s_k\in \Omega_\alpha$ for $k\ge 1$. Then positive homogeneity of $X_{L,\sigma^2}^{(N)}(s)$ implies
\begin{equation}\label{bound_iter}
\|s_m\|=\Bigl\|\underbrace {({\bf X}_{L,\sigma^2}^{(N)} \circ {\bf X}_{L,\sigma^2}^{(N)} \circ \ldots \circ{\bf X}_{L,\sigma^2}^{(N)})}_{m-1 \text{ times}} (s_1)\Bigr\|\le K^{m-1}\|s_1\|,
\end{equation}
which shows that 
\begin{equation}
\lim_{m\to\infty} s_m=0.
\end{equation}
The observation above leads to the following result.

\begin{corollary}\label{cor:basin}
Assume $0 \leq \alpha < 1$,  $\tilde \sigma^2 = 0$ (no bias), and ${\bf X}^{(N)}_{L,\sigma^2}(s)$ has layer widths $n_1=n_2=\ldots=n_{l-1}=n_{l+1}=\ldots=n_{L+1}=N$, $cN\le n_l<N$ for some $l\in[2;L]$ and fixed $c\in (0;1)$. If $\sigma<\beta_\alpha$ defined in~\eqref{def_beta_alpha}, then for $L\geq\widetilde{L}$ with probability $\geq 1 - \widetilde{C} \exp (- \widetilde{c} N)$ (where $\widetilde{L}$, $\widetilde{C}, \widetilde{c} > 0$ depend on $\sigma^2, c$) autoencoder function ${\bf X}_{L,\sigma^2}^{(N)}(s)$ has unique stable fixed point $0$ with basin of attraction $\mathbb{R}^N$.

\end{corollary}

As we see, for autoencoders with ReLU/LeakyReLU activation and no biases global stability of the fixed point is controlled by $\sup\limits_{ s\in\Omega_\alpha,\|s\|=1}\|{\bf X}_{L,\sigma^2}^{(N)}(s)\|$. 
For such autoencoders, we can define \textit{global edge-of-chaos }
as the value $\sigma^2_{L, \text{crit}, \text{glob}}$ for which almost surely
\begin{equation}
\lim_{N\to\infty}\sup_{\substack{ s\in\Omega_\alpha\\\|s\|=1}}\|{\bf X}_{L,\sigma^2}^{(N)}(s)\|= 1.
\end{equation}
Corollary~\ref{cor:aut_L_to_inf} provides a lower bound on global edge-of-chaos in large-depth limit:
\begin{equation}
\liminf\limits_{L\to\infty}\sigma^2_{L,\mathrm{crit,glob}}\ge \beta_\alpha^2.
\end{equation}

Unlike local EoC, which is based on pointwise convergence, global EoC is based on \textbf{uniform} bounds w.r.t.\ the input $s$.

\begin{remark}

One can consider iterations of autoencoder similar to~\eqref{iter_proc}, but with i.i.d.\ copies ${\bf X}_{L_0,\sigma^2}^{(N),(k)}(s)$ of ${\bf X}_{L_0,\sigma^2}^{(N)}(s)$, $k=1,2,\ldots$. This corresponds to the case where on each step of iteration, random weights of autoencoder function are generated anew.  Then 
\begin{equation}
{\bf X}_{L,\sigma^2}^{(N)}(s) = \underbrace {({\bf X}_{L_0,\sigma^2}^{(N),(m)}\circ \ldots \circ {\bf X}_{L_0,\sigma^2}^{(N),(2)}  \circ{\bf X}_{L_0,\sigma^2}^{(N),(1)})}_{m \text{ times}} (s)
\label{composit1}
\end{equation} 
is itself an autoencoder with $L=mL_0$ layers. If we consider the sequence
\begin{equation}
s_m={({\bf X}_{L_0,\sigma^2}^{(N),(m)}\circ \ldots \circ {\bf X}_{L_0,\sigma^2}^{(N),(2)}  \circ{\bf X}_{L_0,\sigma^2}^{(N),(1)})}(s_0),\qquad s_0\in\mathbb{R}^N,
\end{equation}
then 
one can use the results of Theorem~\ref{thm:main_glob} to obtain condition on $\sigma^2$ for which 
\begin{equation}\label{lim_s_0_indep_copies}
\lim_{m\to\infty} s_m=0,
\end{equation}
i.e. ${\bf X}_{L_0,\sigma^2}^{(N)}$ has unique stable fixed point $0$ with basin of attraction $\mathbb{R}^N$ w.r.t.\ iterations of autoencoder with newly generated weights.

In particular, for autoencoder ${\bf X}_{L_0,\sigma^2}^{(N)}$ with layer widths as in Corollary~\ref{cor:basin}, convergence~\eqref{lim_s_0_indep_copies} holds when $\sigma<\beta_\alpha$ for large enough $L_0\geq \widetilde{L}$.

\end{remark}

The paper is organized as follows. Section 2 contains the proof of Theorem~\ref{thm:main_glob} in the ReLU case $\alpha=0$. Section 3 contains the proof of Theorem~\ref{thm:main_loc}.   The proof of Theorem~\ref{thm:main_glob} in the LeakyReLU case $0<\alpha<1$ follows the exact same steps as in Section 2 and is given in Appendix A. 
\bigskip

\textbf{Acknowledgement.} The work of V.S. and R.S. is supported by Grant “International Multilateral Partnerships for Resilient Education and Science System in Ukraine” IMPRESS-U: N7114 funded by US National Academy of Science and Office of Naval Research Global. The work of V.S. is partially supported by NASU project F 26-5. The work of R.S. is partially supported by NASU project M 22-13.
The work of L.B. is partially supported by the NSF Grant IMPRESS-U: N2401227. The work of S.S. and Y.S. is partially supported by the Or programme of the ISF under agreement number 4101/25. The work of Y.S. is partially supported by   by the European Research Council (ERC) under the
European Union’s Horizon 2022 research and innovation program (grant agreement No. 101041711).

\section{Proof of Theorem~\ref{thm:main_glob}, ReLU case $\alpha=0$}\label{sec:thm2}

\subsection{Auxiliary tools}

The main technical tool of the proof is Sudakov--Fernique inequality (\cite{Sud:1971,Sud:1976,Fern:1997}):

\begin{theorem}[Sudakov--Fernique inequality]\label{thm:SF}
Let $\{P_u\}_{u\in S}$ and $\{Q_u\}_{u\in S}$ be centered, almost surely bounded Gaussian processes indexed by a set $S$. 
Assume that for all $u,u'\in S$,
\begin{equation}\label{eq:SF-assump}
\E\bigl(P_u-P_{u'}\bigr)^2\ \le\ \E\bigl(Q_u-Q_{u'}\bigr)^2.
\end{equation}
Then
\[
\E\sup_{u\in S}P_u\ \le\ \E\sup_{u\in S}Q_u.
\]
\end{theorem}

We obtain an upper bound for $\sup\|{\bf X}_{L,\sigma^2}^{(N)}(s)\|$ via the supremum of a certain Gaussian process. The Sudakov--Fernique inequality will be used to estimate the expected supremum of that Gaussian process by another Gaussian process for which expected supremum is computable.

We will also need a version of the Gaussian concentration inequality to show that the supremum of the Gaussian process is close to its expected value with high probability (\cite{Bor:1975, TIS:1975}):

\begin{theorem}[Borell--Tsirelson--Ibragimov--Sudakov (Borell--TIS) inequality]
\label{thm:borell}
Let $\{P_u\}_{u\in S}$ be a centered, almost surely bounded Gaussian process with $M=\sup\limits_{u\in S}\E P_u^2<\infty$.
Then for all $t\ge 0$,
\begin{equation}
\PP\Big\{\sup_{u\in S} P_u - \E\sup_{u\in S} P_u \ge t\Big\}\le \exp\!\left(-\frac{t^2}{2M}\right).
\end{equation}
\end{theorem}

\subsection{Outline of the proof}\label{sbsc:plan_S3}

Consider $L$-layer autoencoder with ReLU activation, no biases, and Gaussian initialization:
\begin{equation}
{\bf X}^{(N)}_{L,\sigma^2}(s)=\left(\Phi^{(L)}\circ \ldots\circ \Phi^{(2)}\circ \Phi^{(1)}\right)(s),\quad \Phi^{(k)}(s)=(W^{(k)} s)_+.
\end{equation}
Here and below we use the notation
\begin{equation}
v_+=((v_1)_+,\ldots,(v_n)_+)\text{\quad for } v\in\mathbb{R}^n
\end{equation}
for brevity.
Also denote
\begin{equation}
\|{\bf X}_{L,\sigma^2}^{(N)}\|_+:=\sup_{\substack{s\in\mathbb{R}^N_+\\ \|s\|=1}} \|{\bf X}_{L,\sigma^2}^{(N)}(s)\|,
\end{equation}
where $\mathbb{R}^N_+$ is defined in~\eqref{def_Omega_alpha}.

We will perform the proof in four steps:

\begin{itemize}

\item
\textbf{Step 1:}
Obtain an upper bound of the form
\begin{equation}
 (\|{\bf X}^{(N)}_{L,\sigma^2}\|_+)^{1/L}\le\dfrac{1}{L}\sup_{u\in S} P_u
\end{equation}
for certain centered Gaussian process $\{P_u\}_{u\in S}$.

\item
\textbf{Step 2:}
Find a centered Gaussian process $\{Q_u\}_{u\in S}$ which majorizes $\{P_u\}_{u\in S}$ in terms of~\eqref{eq:SF-assump} and such that $\E\sup\limits_{u\in S}Q_u$ is computable.

\item
\textbf{Step 3:}
Apply Sudakov-Fernique inequality to obtain an upper bound 
$\dfrac{1}{L}\E\sup\limits_{u\in S}P_u\le \dfrac{\sigma}{\beta_0^{(L)}}$.

\item
\textbf{Step 4:}
Use Borell--TIS inequality to estimate the probability
\begin{equation}
\PP\left\{\dfrac{1}{L}\sup_{u\in S} P_u-\E\Big\{\dfrac{1}{L}\sup_{u\in S} P_u\Big\}>\varepsilon\right\}.
\end{equation}
Then apply such bound to obtain~\eqref{prob_norm<1}. 

\end{itemize}

\subsection{Step 1: Upper bound via Gaussian process}\label{sbsc:thm2_step1}

Fix $s\in\mathbb{R}^N_+$ with $\|s\|=1$. Define neuron vectors and their norms:
\begin{equation}
s^{(0)}:=s\in\mathbb{R}^N_+,\quad 
s^{(k)}:=\Phi^{(k)}(s^{(k-1)})=(W^{(k)} s^{(k-1)})_+\in\mathbb{R}^{n_{k+1}}_+,\quad
r^{(k)}:=\|s^{(k)}\|.
\end{equation}
Then $s^{(L)}={\bf X}^{(N)}_{L,\sigma^2}(s)$ and we can write
\begin{equation}
\|{\bf X}^{(N)}_{L,\sigma^2}(s)\|=r^{(L)}=\prod_{k=1}^L\dfrac{r^{(k)}}{r^{(k-1)}}.
\end{equation}
Thus, we can rewrite $\|{\bf X}^{(N)}_{L,\sigma^2}\|_+$ in the following way:
\begin{equation}\label{eq_XNL_rl}
\|{\bf X}^{(N)}_{L,\sigma^2}\|_+=\sup_{s\in\R^N_+,\ \|s\|=1}\prod_{k=1}^L\dfrac{r^{(k)}}{r^{(k-1)}}.
\end{equation}
Also define normalized neuron vectors $u_k:=\dfrac{s^{(k)}}{r^{(k)}}\in \Spos{n_{k+1}-1}$, where $\Spos{n-1}=\mathbb{R}^n_+\cap\mathbb{S}^{n-1}$. Then we can express norm gain ratios $\dfrac{r^{(k)}}{r^{(k-1)}}$ in terms of $u_k$'s:
\begin{equation}\label{eq_rl_norm}
\dfrac{r^{(k)}}{r^{(k-1)}}=\dfrac{1}{r^{(k-1)}}\|(W^{(k)}s^{(k-1)})_+\|=\dfrac{1}{r^{(k-1)}}\|(W^{(k)}r^{(k-1)}u_{k-1})_+\|=\|(W^{(k)}u_{k-1})_+\|.
\end{equation}
Moreover, $u_k$'s satisfy the recurrence relation
\begin{equation}\label{u_l_rec}
u_0=s,\qquad
u_k=\dfrac{(W^{(k)} u_{k-1})_+}{\|(W^{(k)} u_{k-1})_+\|}.
\end{equation}
It is easy to check that for any vector $v\in\mathbb{R}^n$ we have $(v,v_+)=\|v_+\|^2$. Applying this fact for $v=W^{(k)}u_{k-1}$, we obtain
\begin{multline}\label{eq_norm_scalar}
\|(W^{(k)} u_{k-1})_+\|=\dfrac{\|(W^{(k)} u_{k-1})_+\|^2}{\|(W^{(k)} u_{k-1})_+\|}=\dfrac{(W^{(k)} u_{k-1}, (W^{(k)} u_{k-1})_+)}{\|(W^{(k)} u_{k-1})_+\|}=\\
=\left(W^{(k)} u_{k-1}, \dfrac{(W^{(k)} u_{k-1})_+}{\|(W^{(k)} u_{k-1})_+\|}\right)=(W^{(k)} u_{k-1}, u_k).
\end{multline}
Combining~\eqref{eq_XNL_rl}, \eqref{eq_rl_norm} and \eqref{eq_norm_scalar}, we obtain
\begin{equation}\label{eq_XNL_prod_1}
\|{\bf X}^{(N)}_{L,\sigma^2}\|_+=\sup_{s\in\Spos{N-1}} \prod_{k=1}^L (W^{(k)}u_{k-1},u_k).
\end{equation}

\begin{remark}
If for some $k$, $r^{(k)}=0$, then $\|{\bf X}^{(N)}_{L,\sigma^2}(s)\|=0$ and such $s$ has no impact on the upper bound on $\|X^{(N)}_{L,\sigma^2}\|_+$.
\end{remark}

In order to obtain Gaussian process under the supremum, we first use AM-GM inequality:
\begin{equation}
a_1a_2\ldots a_L\le\left(\dfrac{a_1+a_2+\ldots+a_L}{L}\right)^L \text{\quad for } a_k\ge 0.
\end{equation}
We can take $a_k:=(W^{(k)} u_{k-1},u_k)=\|(W^{(k)} u_{k-1})_+\|\ge 0$ and obtain
\begin{equation}
\|{\bf X}^{(N)}_{L,\sigma^2}\|_+\le \sup_{s\in\Spos{N-1}} 
\left(
\dfrac{\sum\limits_{k=1}^L(W^{(k)}u_{k-1},u_k)}{L}
\right)^L.
\end{equation} 

In formula~\eqref{eq_XNL_prod_1} the sequence of $u_k$'s is defined by a single input $s\in\Spos{N-1}$ via the recurrence~\eqref{u_l_rec}. In order to obtain a computable upper bound, we simply `forget' about the dependencies between $u_k$'s, which can only increase the supremum. This observation gives us an upper bound
\begin{equation}\label{norm<indepWuu}
\|{\bf X}^{(N)}_{L,\sigma^2}\|_+\le 
\Bigg(
\dfrac{1}{L}\sup_{\substack{u_k\in\Spos{n_{k+1}-1},\\k=0,1,\ldots,L}}\quad\sum\limits_{k=1}^L(W^{(k)}u_{k-1},u_k)
\Bigg)^L.
\end{equation} 

\begin{remark}
In the case of 1-layer DNN ($L=1$) the upper bound~\eqref{norm<indepWuu} is exact, i.e.
\begin{equation}
\|{\bf X}^{(N)}_{1,\sigma^2}\|_+=\sup_{u_0,u_1\in \Spos{N-1}} (Wu_0,u_1)
\end{equation}
for $X^{(N)}_{1,\sigma^2}(s)=(Ws)_+$.
One can obtain it using  the fact that $\|v_+\|=\sup\limits_{u\in\Spos{N-1}}(v,u)$ for any $v\in\R^N$.

\end{remark}
Denote
\begin{equation}
P_{u_0,u_1,\ldots,u_L}:=\sum\limits_{k=1}^L(W^{(k)}u_{k-1},u_k).
\end{equation}
Since the matrices $W^{(k)}$ are independent and have independent Gaussian entries, $P_{u_0,u_1,\ldots,u_L}$ is a Gaussian random variable for any choice of $(u_0,u_1,\ldots,u_L)\in\Spos{n_1-1}\times\Spos{n_2-1}\times\ldots\times \Spos{n_{L+1}-1}$. Moreover, $\{P_{u_0,u_1,\ldots,u_L}\}_{(u_0,\ldots,u_L)\in\Spos{n_1-1}\times\ldots\times \Spos{n_{L+1}-1}}$ is an almost surely bounded Gaussian process indexed by the set $\Spos{n_1-1}\times\ldots\times \Spos{n_{L+1}-1}$.

In conclusion, we bounded $\|{\bf X}_{L,\sigma^2}^{(N)}\|_+$ via supremum of certain Gaussian process. In the following steps we will first estimate the expectation of that supremum, and then establish bounds on probability of deviation from expected value. However, even expected value of supremum of such Gaussian process is impossible to compute directly, and that is where Sudakov-Fernique comparison inequality is applied.

\subsection{Step 2: Majorizing Gaussian process}

For the sake of brevity, let us denote ${\bf u}:=(u_0,u_1,\ldots,u_L)$ and ${\bf S}_+:=\Spos{n_1-1}\times\Spos{n_2-1}\times\ldots\times \Spos{n_{L+1}-1}$. We have an upper bound for $\|{\bf X}^{(N)}_{L,\sigma^2}\|_+$:
\begin{equation}
\|X_{L,\sigma^2}^{(N)}\|_+\le 
\Bigg(
\dfrac{1}{L}\sup_{{\bf u}\in{\bf S}_+} P_{\bf u}
\Bigg)^L, 
\qquad
P_{\bf u}=\sum\limits_{k=1}^L(W^{(k)}u_{k-1},u_k).
\end{equation} 
The goal of this step is to obtain an upper bound for $\E\left\{\sup\limits_{{\bf u}\in{\bf S}_+} P_{\bf u}\right\}$ using Sudakov-Fernique inequality. 

Let us construct a majorizing Gaussian process. Consider independent Gaussian vectors $g_0,g_1,\ldots,g_L$ such that each $g_k\in\R^{n_{k+1}}$ has i.i.d.\ entries distributed as $N(0,1)$. Denote
\begin{equation}
Q_{\bf u}=Q_{u_0,u_1,\ldots,u_L}:=\sqrt{\dfrac{\sigma^2}{n_1}}(g_0,u_0)+\sum_{k=1}^{L-1}\sqrt{\dfrac{\sigma^2}{n_k}+\dfrac{\sigma^2}{n_{k+1}}}(g_k,u_k)
+\sqrt{\dfrac{\sigma^2}{n_{L}}}(g_L,u_L).
\end{equation}
One can easily check that $\{Q_{\bf u}\}_{{\bf u}\in{\bf S}_+}$ is an almost surely bounded Gaussian process. In order to check majorization condition~\eqref{eq:SF-assump}, we need to compute $\E\{(P_{\bf u}-P_{\bf u'})^2\}$
and $\E\{(Q_{\bf u}-Q_{\bf u'})^2\}$
for arbitrary ${\bf u}, {\bf u'}\in {\bf S}_+$.

Observe that $P_{\bf u}-P_{\bf u'}$ is a centered random variable, which means that $\E\{(P_{\bf u}-P_{\bf u'})^2\}$ is its variance. Then independence of the matrices $W^{(l)}$ implies that
\begin{multline}\label{EP-P'_1}
\E\{(P_{\bf u}-P_{\bf u'})^2\}=
\sum_{k=1}^L \E\{\bigl((W^{(k)}u_{k-1},u_k)-(W^{(k)}u_{k-1}',u_k')\bigr)^2\}
=\\
=\sum_{l=1}^L\Bigl( \E\{(W^{(k)}u_{k-1},u_k)^2\}-2\E\{(W^{(k)}u_{k-1},u_k)(W^{(k)}u_{k-1}',u_k')\}
+\E\{(W^{(k)}u'_{k-1},u'_k)^2\}\Bigr)
\end{multline}
We can write $W^{(k)}=\sqrt{\dfrac{\sigma^2}{n_k}}G^{(k)}$, where $G^{(k)}\in\mathbb{R}^{n_{k+1}\times n_k}$ has i.i.d.\ entries $N(0,1)$. 
Straightforward computation shows that for a Gaussian matrix $G$ with i.i.d.\ entries $N(0,1)$ and arbitrary deterministic vectors $a,b,c,d$ the following identity holds:
\begin{equation}
\E\{(Ga,b)(Gc,d)\}=(a,c)(b,d),
\end{equation}
which implies
\begin{equation}\label{EP-P'_2}
\E\{(P_{\bf u}-P_{\bf u'})^2\}=
\sum_{k=1}^L \dfrac{\sigma^2}{n_k} (\|u_{k-1}\|^2\|u_k\|^2-2(u_{k-1},u_{k-1}')(u_k,u_k')+\|u_{k-1}'\|^2\|u_k'\|^2)
\end{equation}
Taking into account $\|u_k\|=\|u_k'\|=1$, we obtain final formula for $\E\{(P_{\bf u}-P_{\bf u'})^2\}$:
\begin{equation}
\E\{(P_{\bf u}-P_{\bf u'})^2\}
=2\sum_{k=1}^L \dfrac{\sigma^2}{n_k}\bigl(1-(u_{k-1},u_{k-1}')(u_{k},u_k')\bigr)
\end{equation}
Note that $(u_{k-1},u_{k-1}')\le 1$, $(u_{k},u_k')\le 1$ by Cauchy-Schwarz. A trivial inequality
\begin{equation}
1-xy\le 1-x+1-y\text{\quad for }x,y\le 1
\end{equation}
gives us an upper bound for $\E\{(P_{\bf u}-P_{\bf u'})^2\}$:
\begin{multline}\label{VarP_bound}
\E\{(P_{\bf u}-P_{\bf u'})^2\}
\le 2
\sum_{k=1}^L \dfrac{\sigma^2}{n_l}\Bigl(\bigl(1-(u_{k-1},u_{k-1}')\big)+\bigl(1-(u_{k},u_k')\bigr)\Bigr)=\\
=
2\dfrac{\sigma^2}{n_1}(1-(u_0,u_0'))+2\sum_{k=1}^{L-1}\left(\dfrac{\sigma^2}{n_k}+\dfrac{\sigma^2}{n_{k+1}}\right)\bigl(1-(u_k,u_k')\bigr)
+2\dfrac{\sigma^2}{n_{L}}\bigl(1-(u_L,u_L')\bigr)
\end{multline}
Let us move on to $\E\{(Q_{\bf u}-Q_{\bf u'})^2\}$. Due to the independence of $g_k$ we obtain
\begin{multline}
\E\{(Q_{\bf u}-Q_{\bf u'})^2\}=
\dfrac{\sigma^2}{n_1}\E\{(g_0,u_0-u_0')^2\}+\\
+\sum_{k=1}^{L-1}\left(\dfrac{\sigma^2}{n_k}+\dfrac{\sigma^2}{n_{k+1}}\right) \E\{\bigl(g_k,u_k-u_k'\bigr)^2\}
+\dfrac{\sigma^2}{n_{L}}\E\{(g_L,u_L-u_L')^2\}.
\end{multline}
For a Gaussian vector $g$ with i.i.d.\ entries $N(0,1)$ and an arbitrary deterministic vector $v$ one have
\begin{equation}\label{E(g,v)^2}
\E\{(g,v)^2\}=\|v\|^2.
\end{equation}
Recall that $\|u_k\|=\|u_k'\|=1$, which means that $\|u_k-u_k'\|^2=2(1-(u_k,u_k'))$. Combining this with~\eqref{E(g,v)^2}, we obtain
\begin{equation}\label{VarQ_formula}
\E\{(Q_{\bf u}-Q_{\bf u'})^2\}
= 
2\dfrac{\sigma^2}{n_1}(1-(u_0,u_0'))+
2\sum_{k=1}^{L-1}\left(\dfrac{\sigma^2}{n_k}+\dfrac{\sigma^2}{n_{k+1}}\right)\bigl(1-(u_k,u_k')\bigr)
+2\dfrac{\sigma^2}{n_{L}}\bigl(1-(u_L,u_L')\bigr)
\end{equation}
The upper bound~\eqref{VarP_bound} and the identity~\eqref{VarQ_formula} together yield that
\begin{equation}
\E\{(P_{\bf u}-P_{\bf u'})^2\}\le \E\{(Q_{\bf u}-Q_{\bf u'})^2\},
\end{equation}
i.e.\ Gaussian process $Q_{\bf u}$ majorizes $P_{\bf u}$ in terms of~\eqref{eq:SF-assump}.

\subsection{Step 3: Upper bound on expected supremum}
\label{sbsc:thm2_step3}

In Step 2 we checked that Gaussian processes $P_{\bf u}$ and $Q_{\bf u}$ satisfy condition~\eqref{eq:SF-assump}.
Then Sudakov-Fernique inequality yields that
\begin{equation}\label{EsupP<EsupQ}
\E\{\sup_{{\bf u}\in{\bf S}_+} P_{\bf u}\}\le \E\{\sup_{{\bf u}\in{\bf S}_+} Q_{\bf u}\}. 
\end{equation}
We are left to compute $\E\{\sup\limits_{{\bf u}\in{\bf S}_+} Q_{\bf u}\}$ to complete the upper bound for $\E\{\sup\limits_{{\bf u}\in{\bf S}_+} P_{\bf u}\}$. Observe that for any vector $v\in\mathbb{R}^n$ one can compute the following supremum:
\begin{equation}
\sup_{u\in S^{n-1}_+} (v,u)=\|v_+\|.
\end{equation}
This formula together with the definition of $Q_{\bf u}$ gives us
\begin{equation}\label{sup_Q_|g|}
\E\{\sup\limits_{{\bf u}\in{\bf S}_+} Q_{\bf u}\}=
\sqrt{\dfrac{\sigma^2}{n_1}}\E\{\|(g_0)_+\|\}+\sum_{k=1}^{L-1}\sqrt{\dfrac{\sigma^2}{n_k}+\dfrac{\sigma^2}{n_{k+1}}}\E\{\|(g_k)_+\|\}
+\sqrt{\dfrac{\sigma^2}{n_{L}}}\E\{\|(g_L)_+\|\}
\end{equation}
For a Gaussian vector $g=(g^{(1)},\ldots,g^{(n)})\in\mathbb{R}^n$ with i.i.d.\ entries $N(0,1)$ one can use Cauchy-Schwarz inequality to obtain
\begin{equation}
\E\{\|g_+\|\}\le (\E\{\|g_+\|^2\})^{1/2}=\left(\sum_{j=1}^n \E\{(g^{(j)}_+)^2\}\right)^{1/2}=\sqrt{\dfrac{n}{2}}.
\end{equation} 
Using this bound for $g_0,g_1,\ldots,g_L$ and recalling that $n_1=N$, $n_{L+1}=N$, $n_k=c_k(N)\cdot N$, we obtain that \eqref{sup_Q_|g|} together with~\eqref{EsupP<EsupQ} implies
\begin{equation}\label{supP_K}
\dfrac{1}{L}\E\{\sup\limits_{{\bf u}\in{\bf S}_+} P_{\bf u}\}\le 
\sigma\cdot\dfrac{1+\sum\limits_{k=1}^{L-1}\sqrt{1+\dfrac{c_{k+1}}{c_k}}+\sqrt{\dfrac{1}{c_L}}}{\sqrt{2}\,L}=\dfrac{\sigma}{\beta_0^{(L)}}.
\end{equation}

\begin{remark}
Despite the fact that we used Cauchy-Schwarz to obtain only an upper bound on $\E\{\|g_+\|\}$, such bound is in fact asymptotically sharp as $n\to\infty$, i.e. $\E\{\|g_+\|\}\asymp \sqrt{\dfrac{n}{2}}$. Thus, there is no significantly better bound on $\dfrac{1}{L}\E\{\sup\limits_{{\bf u}\in{\bf S}_+} P_{\bf u}\}$ if we use this particular $Q_{\bf u}$ as a majorizing Gaussian process.
\end{remark}

\subsection{Step 4: Gaussian concentration}\label{sbsc:thm2_step4}

We are left to prove that there exists $\widetilde{c}>0$ such that
\begin{equation}\label{supP_concentr}
\PP\left\{\dfrac{1}{L}\sup\limits_{{\bf u}\in{\bf S}_+} P_{\bf u}-\E\Bigl\{\dfrac{1}{L}\sup\limits_{{\bf u}\in{\bf S}_+} P_{\bf u}\Bigr\}\ge\varepsilon\right\}\le e^{-\widetilde{c} NL}.
\end{equation}

We will use Borell--TIS inequality (Theorem~\ref{thm:borell}) to obtain~\eqref{supP_concentr}. To this end we need to estimate $\E\{P^2_{\bf u}\}$. Similarly to~\eqref{EP-P'_1} and \eqref{EP-P'_2} we obtain
\begin{equation}
\E\{P^2_{\bf u}\}=
\sum_{k=1}^L \E\{\bigl(W^{(k)}u_{k-1},u_k\bigr)^2\}
=
\sum_{k=1}^L \dfrac{\sigma^2}{n_k} \|u_{k-1}\|^2\|u_k\|^2=
\dfrac{S}{N}
\text{\quad for }S=\sum_{k=1}^L \dfrac{\sigma^2}{c_k}\le L\dfrac{\sigma^2}{c}.
\end{equation}
Thus, in terms of Theorem~\ref{thm:borell} $M=\sup\limits_{{\bf u}\in{\bf S}_+} \E\{P^2_{\bf u}\}\le\dfrac{L\sigma^2}{Nc}$. Applying Theorem~\ref{thm:borell} for $t=L\varepsilon$, we obtain
\begin{equation}
\PP\Big(\sup_{{\bf u}\in {\bf S}_+} P_{\bf u} - \E\sup_{{\bf u}\in {\bf S}_+} P_{\bf u} \ge L\varepsilon\Big)\le \exp\!\left(-\frac{NL\varepsilon^2c}{2\sigma^2}\right),
\end{equation}
which gives us~\eqref{supP_concentr} with $\widetilde{c}=\dfrac{\varepsilon^2c}{2\sigma^2}$ and completes the proof of Theorem~\ref{thm:main_glob} for $\alpha=0$.

\section{Proof of Theorem~\ref{thm:main_loc}}

Consider input-output Jacobian ${\bf J}_{L,\sigma^2}^{(N)}(s)$ \eqref{IO_def} of autoencoder ${\bf X}^{(N)}_{L,\sigma^2}(s)$ with ReLU activation, Gaussian initialization, and asymptotically proportional layers:
\begin{equation}
\dfrac{n_k}{N}\to c_k\in(c;C)\quad\text{as $n\to\infty$, $k=2,3,\ldots,L$}.
\end{equation}
Since $n_1=n_{L+1}=N$, we also denote $c_1=c_{L+1}=1$ for convenience.
Since random matrix ${\bf J}_{L,\sigma^2}^{(N)}(s)$ is generally non-symmetric, its norm is determined by its singular values, i.e. by the eigenvalues of ${\bf M}_{L,\sigma^2}^{(N)}(s)=({\bf J}_{L,\sigma^2}^{(N)}(s))^T{\bf J}_{L,\sigma^2}^{(N)}(s)$. Denote $\mu_+^{(N)}(s)$ the maximal eigenvalue of ${\bf M}_{L,\sigma^2}^{(N)}(s)$, then 
\begin{equation}
\|{\bf J}_{L,\sigma^2}^{(N)}(s)\|=\sqrt{\mu_+^{(N)}(s)}
\end{equation}
We are interested in the behavior of $\mu_+^{(N)}(s)$ in the double limit $\lim\limits_{L\to\infty}\lim\limits_{N\to\infty}$.
It was proved in \cite{Pa:2020} and \cite{Pa-Sl:2023} that for a given sequence of input vectors $s_N\in\mathbb{R}^N$  the empirical spectral distribution of ${\bf M}_{L,\sigma^2}^{(N)}(s_N)$ converges to a non-random probability measure $\nu_{\scriptscriptstyle{\bf M}_{L,\sigma^2}}$. 

 Theorem~\ref{thm:main_loc} will follow from the following two facts:

\begin{proposition}[Right edge of limiting eigenvalue distribution]\label{pro:sect2_1}
Let $\mu_+^{(\infty)}$ be the right edge of  the limiting eigenvalue distribution $\text{supp}\,\nu_{\scriptscriptstyle{\bf M}_{L,\sigma^2}}$ of ${\bf M}_{L,\sigma^2}^{(N)}(s_N)$. Then:
\begin{enumerate}
\item[(i)] $\mu_+^{(\infty)}$ does not depend on the choice of $s_N$ and 
\begin{equation}\label{prosect21:right_edge}
\mu_+^{(\infty)}=\left(\dfrac{\sigma^2}{2}\right)^L \cdot m_*^{-1}(m_*+1)\cdot\prod\limits_{k=1}^L (2c_k m_*+1),\quad \text{where } m_*\asymp \dfrac{1}{L};
\end{equation}

\item[(ii)] 
In the infinite-width limit:
\begin{equation}\label{eq_mu_+_lim}
\lim_{L\to\infty}\dfrac{1}{L}\log\mu_+^{(\infty)}
=\log(\sigma^2/2);
\end{equation}

\item[(iii)] Let $\sigma_{L,\mathrm{crit}}^2$ be the value of $\sigma^2$ for which $\mu_+^{(\infty)}=1$. Then in the infinite-depth limit:
\begin{equation}\label{eq_sigma_crit_lim}
\lim_{L\to\infty} \sigma_{L,\mathrm{crit}}^2=2.
\end{equation}

\end{enumerate}

\end{proposition}

\begin{proposition}[Convergence of spectral norm]\label{pro:seq2_2}
Let $\{s_N\}_{N=1}^\infty$ be an arbitrary sequence of vectors such that $s_N\in\mathbb{R}^N$. Then 
\begin{equation}
\|{\bf J}_{L,\sigma^2}^{(N)}(s_N)\|\to \sqrt{\mu_+^{(\infty)}}\quad \text{a.s. as $N\to\infty$}.
\end{equation}
\end{proposition}
\bigskip

\begin{remark}
In \cite[Section 2.5.1]{Pe-Ga:2017} it was shown that in the infinite-width limit:
\begin{equation}
\lim_{N\to\infty}\|{\bf J}_{L,\sigma^2}^{(N)}(s_N)\|^2=\left(\dfrac{\sigma^2}{2}\right)^L(2eL+O(1))
\end{equation}
in the case when all the layers have the same width, i.e.\ $n_k=N$ for all $k$. The result was obtained by deriving the asymptotic of $\mu_+^{(\infty)}$. Statement (i) of Proposition~\ref{pro:sect2_1} generalizes the result of \cite{Pe-Ga:2017} for arbitrary proportional widths of layers, while the proof of Proposition~\ref{pro:sect2_1} provides a rigorous derivation of \eqref{prosect21:right_edge}. Proposition~\ref{pro:seq2_2} justifies the use of right edge $\mu_+^{(\infty)}$ as infinite-width limit of $\|{\bf J}_{L,\sigma^2}^{(N)}(s_N)\|^2$ in~\cite{Pe-Ga:2017}.
\end{remark}

\subsection{Proof of Proposition~\ref{pro:sect2_1}}

\subsubsection{Outline of the proof}\label{sbsc:sec2_outline}

We will perform the proof in three steps:
\begin{itemize}

\item
\textbf{Step 1:}
Obtain a functional equation on the \textit{moment generating function} of the eigenvalue distribution of ${\bf M}_{L,\sigma^2}^{(N)}(s)$ in infinite-width limit.

\item
\textbf{Step 2:}
Derive formula~\eqref{prosect21:right_edge} right the edge  $\mu_+^{(\infty)}$ of spectrum of ${\bf M}_{L,\sigma^2}^{(N)}(s)$ 
using the equation obtained in \textbf{Step 1}.

\item
\textbf{Step 3:}
Obtain the limits \eqref{eq_mu_+_lim} and \eqref{eq_sigma_crit_lim} using~\eqref{prosect21:right_edge}.

\end{itemize}

The equation mentioned in \textbf{Step 1} was earlier obtained in \cite{Pe-Ga:2018} in the partial case of DNN with layers of the same width and under certain simplifications. Namely, one can use chain rule to represent I/O Jacobian in a form
\begin{equation}\label{JDW}
{\bf J}^{(N)}_{L,\sigma^2}(s)=D^{(L)} W^{(L)}\ldots D^{(2)} W^{(2)} D^{(1)}W^{(1)},
\end{equation}
where $D^{(k)}$ are $n_{k+1}\times n_{k+1}$ diagonal matrices having $\lambda^\prime((W^{(k)}s^{(k-1)})_j
+b^{(k)}_j
)$ on diagonals. In \cite{Pe-Ga:2018} matrices $W^{(k)}$ and $D^{(j)}$ were treated as independent. In the case of ReLU activation one can consider a matrix
\begin{equation}\label{JDW_tilde}
\widetilde{{\bf J}}^{(N)}_{L,\sigma^2}=\widetilde D^{(L)} W^{(L)}\ldots \widetilde D^{(2)} W^{(2)} \widetilde D^{(1)}W^{(1)},
\end{equation}
where $\widetilde D^{(k)}$ are $n_{k+1}\times n_{k+1}$ diagonal matrices having i.i.d.\  $\mathrm{Bernoulli}\,(\frac{1}{2})$ variables independent of $W^{(j)}$ on diagonals. The results of \cite{Pa:2020,Pa-Sl:2023} imply that in ReLU activation case,  in the infinite-width limit the densities of singular value distribution of ${\bf J}^{(N)}_{L,\sigma^2}(s)$ and $\widetilde{{\bf J}}^{(N)}_{L,\sigma^2}$ are the same (e.g., \cite[Theorem 2.5]{Pa-Sl:2023}).
In \cite[Proposition 2]{Ha-Ni:2018} it was proved that random matrices ${\bf M}^{(N)}_{L,\sigma^2}(s)=({\bf J}^{(N)}_{L,\sigma^2}(s))^T{\bf J}^{(N)}_{L,\sigma^2}(s)$ and $\widetilde{{\bf M}}^{(N)}_{L,\sigma^2}=(\widetilde{{\bf J}}^{(N)}_{L,\sigma^2})^T\widetilde{{\bf J}}^{(N)}_{L,\sigma^2}$ are equal in distribution for any $s\in\mathbb{R}^N$. Both these results justify the independence assumption used in~\cite{Pe-Ga:2018}.

It was also mentioned in \cite{Pa-Sl:2023} that the results of the paper can be used to generalize the equation on moment generating function in~\cite{Pe-Ga:2018} for the case of the layers of proportional widths.   Here in \textbf{Step 1} we finally derive generalized version of such equation in a closed form (for proportional layers and ReLU activation).

\textbf{Step 2} is the main technical step of the proof. In case of ReLU networks, in \cite{Pe-Ga:2018} the first two moments of eigenvalue distribution of $\widetilde{{\bf M}}^{(N)}_{L,\sigma^2}$ were derived in order to study classical EoC for DNNs. 
 Dealing with the right edge of spectrum requires 
 more subtle analysis. In \cite{Pe-Ga:2017} an asymptotic formula for $\mu_+^{(\infty)}$ (as $L\to\infty$) was derived in the case of layers of the same width. Here in step 2 we obtain an asymptotic formula for  $\mu_+^{(\infty)}$ for arbitrary proportional widths of layers. Moreover, we derive an exact expression for $\mu_+^{(\infty)}$ for the layers of the same width (see Remark~\ref{rem:right_edge_exact}).
The main complication consists in the fact that the moment generating function of the eigenvalue distribution can be found only implicitly.

In \textbf{Step 3} we use the derived formula for the right edge $\mu_+^{(\infty)}$ to obtain local EoC in the limit $L\to\infty$.

\subsubsection{Step 1: Equation on moment generating function.}

We start by introducing some standard notions from Random Matrix Theory concerning eigenvalue distributions. Consider a sequence of random $N\times N$ real symmetric matrices ${\bf K}^{(N)}$. Denote $\{\lambda^{(N)}_k\}_{k=1}^N$ the eigenvalues of ${\bf K}^{(N)}$. \textit{Normalized counting measure (NCM)} of ${\bf K}^{(N)}$ is a random measure defined by
\begin{equation}
\nu_{\scriptscriptstyle{\bf K}^{(N)}}=\dfrac{1}{N}\sum_{k=1}^N \delta_{\scriptscriptstyle\lambda^{(N)}_k}.
\end{equation}
If $\nu_{\scriptscriptstyle{\bf K}^{(N)}}$ converges almost surely to a certain non-random measure $\nu_{\scriptscriptstyle\bf K}$ in a weak sense, we call   $\nu_{\scriptscriptstyle\bf K}$ \textit{limiting NCM} of ${\bf K}^{(N)}$.

\textit{Moment generating function} of ${\bf K}^{(N)}$ is 
\begin{equation}
m_{\scriptscriptstyle{\bf K}}(z)=\sum_{k=1}^\infty a_kz^k,
\end{equation}
where $a_k$ is $k$-th moment of the measure $\nu_{\scriptscriptstyle\bf K}$:
\begin{equation}
a_k=\int_{-\infty}^\infty \lambda^k\, d\nu_{\scriptscriptstyle\bf K}(\lambda).
\end{equation} 
Also denote $z_{\scriptscriptstyle{\bf K}}(m)$ the functional inverse of $m_{\scriptscriptstyle{\bf K}}(z)$.

Recall that ${\bf M}^{(N)}_{L,\sigma^2}(s)=({\bf J}_{L,\sigma^2}^{(N)}(s))^T{\bf J}^{(N)}_{L,\sigma^2}(s)$, where ${\bf J}^{(N)}_{L,\sigma^2}(s)$ is I/O Jacobian of autoencoder function $X_{L,\sigma^2}^{(N)}(s)$.
It was shown in \cite{Pa:2020} that NCM $\nu_{\scriptscriptstyle{\bf M}^{(N)}_{L,\sigma^2}}$ of ${\bf M}^{(N)}_{L,\sigma^2}(s)$ converges almost surely to certain a non-random measure $\nu_{\scriptscriptstyle{\bf M}_{L,\sigma^2}}$, i.e.\ the limiting NCM $\nu_{\scriptscriptstyle{\bf M}_{L,\sigma^2}}$ of ${\bf M}^{(N)}_{L,\sigma^2}(s)$ exists.

The equation on $m_{\scriptscriptstyle{\bf M}_{L,\sigma^2}}$  when ${\bf X}_{L,\sigma^2}^{(N)}(s)$ is a DNN with square layer matrices ($n_1=n_2=\ldots=n_{L+1}$) was obtained in \cite{Pe-Ga:2018} under the conjecture of independence of $W^{(k)}$ and $D^{(j)}$ (see \eqref{JDW} and \eqref{JDW_tilde}). It was later proven in \cite{Pa-Sl:2023} that we can consider ${\bf J}^{(N)}_{L,\sigma^2}$ defined as in~\eqref{JDW} but with certain $D^{(j)}$ independent of $W^{(k)}$, and the eigenvalue distribution of ${\bf M}^{(N)}_{L,\sigma^2}(s)=({\bf J}_{L,\sigma^2}^{(N)}(s))^T{\bf J}^{(N)}_{L,\sigma^2}(s)$ in infinite-width limit remains the same, which justifies the result of \cite{Pe-Ga:2018}. In the case of ReLU activation, after such replacement we obtain  $\widetilde{{\bf M}}^{(N)}_{L,\sigma^2}=(\widetilde{{\bf J}}^{(N)}_{L,\sigma^2})^T\widetilde{{\bf J}}^{(N)}_{L,\sigma^2}$ with $\widetilde{{\bf J}}^{(N)}_{L,\sigma^2}$ defined in \eqref{JDW_tilde}, i.e. $\nu_{\scriptscriptstyle{\bf M}_{L,\sigma^2}}=\nu_{\scriptscriptstyle{\bf \widetilde{M}}_{L,\sigma^2}}$.

\begin{remark}
Once we replaced ${\bf J}^{(N)}_{L,\sigma^2}(s)$ by $\widetilde{{\bf J}}^{(N)}_{L,\sigma^2}$, the measure $\nu_{\scriptscriptstyle{\bf M}_{L,\sigma^2}}(z)$ and consequently $\mu_+^{(\infty)}$ does not depend on the choice of $s$.
\end{remark}

Moreover, \cite{Pa-Sl:2023} considered the case of DNNs with rectangular layer matrices of proportional sizes rather than square ones, with $\sigma^2=1$, and derived a recurrent relation on $m_{\scriptscriptstyle{\bf M}_{L,\sigma^2}}$. We will use a slightly modified version of that relation for arbitrary value of weight variance parameter $\sigma^2$. Let us formulate that relation in a shortest way.

One can easily check that all matrices $(\widetilde{D}^{(k)})^2$ from~\eqref{JDW_tilde} have limiting NCM
\begin{equation}
\nu_{\scriptscriptstyle{\bf \widetilde D}^{2}}=\dfrac{1}{2}(\delta_0+\delta_1),
\end{equation}
where $\delta_x$ is the delta-measure at $x$.
Let $m_{\scriptscriptstyle{\bf \widetilde D}^{2}}(z)$ be the moment generating function of $\nu_{\scriptscriptstyle{\bf \widetilde D}^{2}}$. Assume that the sizes of weight matrices satisfy $\dfrac{n_k}{n_{k+1}}\to d_k\in(0,+\infty)$.   Then  \cite[Corollary 2.3 and Remark 2.4]{Pa-Sl:2023} imply that for $\sigma^2=1$ the functional inverse $z_{\scriptscriptstyle{\bf M}_{L,1}}(m)$ of $m_{\scriptscriptstyle{\bf M}_{L,1}}(z)$ satisfies
\begin{equation}\label{z_rec_old}
z_{\scriptscriptstyle{\bf M}_{L,1}}(m)=
z_{\scriptscriptstyle{\bf M}_{L-1,1}}(d_Lm)\cdot
z_{\scriptscriptstyle{\bf \widetilde D}^{2}}(d_Lm)\cdot m^{-1}.
\end{equation}
Observe that changing the variance of weights from $\dfrac{1}{n_k}$ to $\dfrac{\sigma^2}{n_k}$ in $\widetilde{{\bf J}}^{(N)}_{L,\sigma^2}$ is equivalent to multiplying the matrices $W^{(k)}$ by $\sigma$, which means $\widetilde{{\bf M}}_{L,\sigma^2}= (\sigma^2)^L \widetilde{{\bf M}}_{L,1}$ and $z_{\scriptscriptstyle{\bf M}_{L,\sigma^2}}(m)=(\sigma^2)^L z_{\scriptscriptstyle{\bf M}_{L,1}}(m)$. Thus, for arbitrary variance parameter $\sigma^2$ relation~\eqref{z_rec_old} becomes
\begin{equation}\label{z_rec_2}
z_{\scriptscriptstyle{\bf M}_{L,\sigma^2}}(m)=
z_{\scriptscriptstyle{\bf M}_{L-1,\sigma^2}}(d_Lm)\cdot
z_{\scriptscriptstyle{\bf \widetilde D}^{2}}(d_Lm)\cdot(\sigma^2m)^{-1}.
\end{equation}
Direct computation shows that
\begin{equation}\label{m_D}
m_{\scriptscriptstyle{\bf D}^{2}}(z)=
\dfrac{1}{2}\,\dfrac{z}{1-z},
\qquad
z_{\scriptscriptstyle{\bf D}^{2}}(m)
=\dfrac{2m}{2m+1}.
\end{equation}
Substituting \eqref{m_D} into \eqref{z_rec_2} gives us
\begin{equation}\label{z_rec_3}
z_{\scriptscriptstyle{\bf M}_{L,\sigma^2}}(m)=
\dfrac{2}{\sigma^2}\,
\dfrac{z_{\scriptscriptstyle{\bf M}_{L-1,\sigma^2}}(d_Lm)}{2d_Lm+1}
\end{equation}
for ReLU activation.
Here we set ${\bf M}_{0,\sigma^2}^{(N)}=I^{(N)}$, which implies that $m_{\scriptscriptstyle{\bf M}_{0,\sigma^2}}(z)=\dfrac{z}{1-z}$, $z_{\scriptscriptstyle{\bf M}_{0,\sigma^2}}(m)=\dfrac{m}{m+1}$. Also recall that the autoencoder's layer sizes $n_1=n_{L+1}=N$ and $n_k=c_kN$, thus $d_k=\dfrac{c_k}{c_{k+1}}$ (here and below $c_1=c_{L+1}=1$). Putting everything together, we obtain an explicit formula for $z_{\scriptscriptstyle{\bf M}_{L,\sigma^2}}(m)$ for ReLU autoencoders:
\begin{equation}\label{z_entier_1}
z_{\scriptscriptstyle{\bf M}_{L,\sigma^2}}(m)
=\left(\dfrac{2}{\sigma^2}\right)^L
\dfrac{m}{(m+1)\cdot\prod\limits_{k=1}^L (2c_k m+1)}.
\end{equation}
Finally, plug $m=m_{\scriptscriptstyle{\bf M}_{L,\sigma^2}}(z)$ into \eqref{z_entier_1} and use $z_{\scriptscriptstyle{\bf M}_{L,\sigma^2}}(m_{\scriptscriptstyle{\bf M}_{L,\sigma^2}}(z))=z$, then we obtain an equation on $m_{\scriptscriptstyle{\bf M}_{L,\sigma^2}}(z)$:
\begin{equation}\label{m_z_eq}
z=\left(\dfrac{2}{\sigma^2}\right)^L
\dfrac{m_{\scriptscriptstyle{\bf M}_{L,\sigma^2}}(z)}{(m_{\scriptscriptstyle{\bf M}_{L,\sigma^2}}(z)+1)\cdot\prod\limits_{k=1}^L (2c_k m_{\scriptscriptstyle{\bf M}_{L,\sigma^2}}(z)+1)}
\end{equation}
The derivation of \eqref{m_z_eq} completes \textbf{Step 1} of the proof.

\subsubsection{Step 2: Formula for the right edge of spectrum}

We start with rewriting equation~\eqref{m_z_eq} in the form
\begin{equation}\label{m_z_eq_F}
\mathcal{F}(m_{\scriptscriptstyle {\bf M}_{L,\sigma^2}}(z),z)=0,
\end{equation}
where
\begin{equation}
\mathcal{F}(m,z)=2^L m-z(\sigma^2)^L\cdot (m+1)\cdot\prod\limits_{k=1}^L (2c_k m+1).
\end{equation}
Such equation cannot be solved with respect to $m_{\scriptscriptstyle {\bf M}_{L,\sigma^2}}(z)$ explicitly for $L>2$. That is the reason why we need to extract the information about the right edge  of spectrum of ${\bf M}_{L,\sigma^2}^{(N)}$ from the equation~\eqref{m_z_eq_F} rather than from explicit formula for $m_{\scriptscriptstyle {\bf M}_{L,\sigma^2}}(z)$.

Recall that $\mu_+^{(\infty)}$ is the right edge of $\text{supp}\, \nu_{\scriptscriptstyle {\bf M}_{L,\sigma^2}}$. Notice that $\text{supp}\, \nu_{\scriptscriptstyle {\bf M}_{L,\sigma^2}}\subset [0;+\infty)$ since ${\bf M}_{L,\sigma^2}^{(N)}$ is a nonnegative matrix.  
A simple argument provides a formula for the right edge of compactly supported measure on $\mathbb{R}_+$ in terms of its moments:
\begin{equation}\label{right_edge}
\mu_+^{(\infty)}=\limsup_{k\to\infty} a_k^{1/k},\qquad
a_k=\int_{-\infty}^{\infty} \lambda^k\,d\nu_{\scriptscriptstyle {\bf M}_{L,\sigma^2}}(\lambda).
\end{equation} 
At the same time, $a_k$ are the coefficients of power series $m_{\scriptscriptstyle {\bf M}_{L,\sigma^2}}(z)=\sum\limits_{k=1}^\infty a_kz^k$ which has radius of convergence $R$ given by
Cauchy-Hadamard theorem:
\begin{equation}\label{C-H}
\dfrac{1}{R}=\limsup_{k\to\infty} a_k^{1/k}.
\end{equation}
As we see, \eqref{right_edge} and \eqref{C-H} yield
\begin{equation}\label{right_edge_R}
\mu_+^{(\infty)}=\dfrac{1}{R}.
\end{equation}
We are left to find the radius of convergence $R$ of the analytic function $m_{\scriptscriptstyle {\bf M}_{L,\sigma^2}}(z)$ given implicitly by \eqref{m_z_eq_F}. A rather standard argument on singularities of implicitly given analytic function and their relation to the radius of convergence shows that there exist $m_*>0$ such that $z=R$, $m=m_*$ is a solution of the system
\begin{equation}\label{system_R_conv}
\begin{cases}
\mathcal{F}(m,z)=0;\\
\dfrac{\partial\mathcal{F}}{\partial m}(m,z)=0.
\end{cases}
\end{equation}
A rigorous argument is placed in Appendix B.

Observe that
\begin{equation}
\dfrac{\partial\mathcal{F}}{\partial m}(m,z)=2^L-z(\sigma^2)^L (m+1)\prod\limits_{k=1}^L (2c_k m+1)\cdot\left(\dfrac{1}{m+1}+\sum_{k=1}^L\dfrac{2c_k}{2c_k m+1}\right).
\end{equation}
Straightforward transformations show that system~\eqref{system_R_conv} implies
\begin{equation}
\dfrac{1}{m+1}+\sum_{k=1}^L\dfrac{2c_k}{2c_k m+1}=\dfrac{1}{m},
\end{equation}
which is equivalent to 
\begin{equation}\label{eq_m_*}
\dfrac{1}{1+\dfrac{1}{m}}+\sum_{k=1}^L\dfrac{1}{1+\dfrac{1}{2c_km}}=1.
\end{equation}
Denote $\widetilde{c}=\min\{1,2c\}$, $\widetilde{C}=\max\{2,2C\}$, then $\widetilde{c}m<m<\widetilde{C}m$ and $\widetilde{c}m<2c_km<\widetilde{C}m$ for $k=1,2,\ldots, L$. Using such estimations and monotonicity of~\eqref{eq_m_*} one can show that the positive solution $m_*$ of \eqref{eq_m_*} is unique and satisfies
\begin{equation}
\dfrac{1}{\widetilde{C}L}<m_*<\dfrac{1}{\widetilde{c}L}.
\end{equation}
Substituting $m=m_*$ into the first equation of \eqref{system_R_conv}, we obtain the solution $(m_*,R)$ of \eqref{system_R_conv} with
\begin{equation}
R=\left(\dfrac{2}{\sigma^2}\right)^L \dfrac{m_*}{(m_*+1)\cdot\prod\limits_{k=1}^L (2c_k m_*+1)}.
\end{equation}
Using \eqref{right_edge_R} we can conclude that
\begin{equation}\label{right_edge_fin}
\mu_+^{(\infty)}=\left(\dfrac{\sigma^2}{2}\right)^L \cdot m_*^{-1}(m_*+1)\cdot\prod\limits_{k=1}^L (2c_k m_*+1),\quad \text{where } \dfrac{1}{\widetilde{C}L}<m_*<\dfrac{1}{\widetilde{c}L},
\end{equation}
which completes the proof of \textit{(i)} and \textbf{Step 2}.

\begin{remark}\label{rem:right_edge_exact}
In case of autoencoder ${\bf X}_{L,\sigma^2}^{(N)}$ with layers of the same size ($n_1=n_2=\ldots=n_{L+1}=N$, i.e. $c_k=1$), the equation~\eqref{eq_m_*} can be solved explicitly and the exact value of the right edge $\mu_+^{(\infty)}$ can be found (see \cite{Sla:2026} for the details):
\begin{equation}
\mu_+^{(\infty)}=\left(\dfrac{\sigma^2}{2}\right)^L\dfrac{(\sqrt{L^2+1}+1)^L}{L^L}\,\frac{\sqrt{L^2+1}+1+L}{\sqrt{L^2+1}+1-L}
\end{equation}

\end{remark}

\subsubsection{Step 3: right edge and $\sigma^2_{crit}(L)$ in the infinite-width limit}

In order to obtain~\eqref{eq_mu_+_lim}, rewrite~\eqref{right_edge_fin} as
\begin{equation}\label{eq_log_mu_long}
\dfrac{1}{L}\log \mu_+^{(\infty)}=
\log \dfrac{\sigma^2}{2}-\dfrac{1}{L}\log m_*+\dfrac{1}{L}\log (1+m_*)+\dfrac{1}{L}\sum_{k=1}^L \log (1+2c_km_*).
\end{equation}
Since $m_*\asymp \dfrac{1}{L}$ and
\begin{equation}
0<\dfrac{1}{L}\sum_{k=1}^L \log (1+2c_km_*)\le \log (1+\widetilde{C}m_*),
\end{equation}
all the summands of the right hand side of~\eqref{eq_log_mu_long} apart from the first one converge to $0$ as $L\to\infty$, which gives us~\eqref{eq_mu_+_lim}. And completes the proof of \textit{(ii)}.

In order to prove~\eqref{eq_sigma_crit_lim}, we first use~\eqref{right_edge_fin} to solve $\mu_+^{(\infty)}=1$ with respect to $\sigma^2$:
\begin{equation}
\sigma_{L,\mathrm{crit}}^2=
2\cdot \Biggl(\dfrac{m_*}{(m_*+1)\cdot\prod\limits_{k=1}^L (2c_k m_*+1)}\Biggr)^{1/L},\quad \text{where } \dfrac{1}{\widetilde{C}L}<m_*<\dfrac{1}{\widetilde{c}L}.
\end{equation}
Rewrite the formula as
\begin{equation}\label{log_sigma}
\log \dfrac{\sigma^2_{crit}(L)}{2}=\dfrac{1}{L}\log m_*-\dfrac{1}{L}\log (1+m_*)-\dfrac{1}{L}\sum_{k=1}^L \log (1+2c_km_*).
\end{equation}
As we already showed, the right hand side converges to zero as $L\to\infty$, which gives us~\eqref{eq_sigma_crit_lim} and completes the prooof of Proposition~\ref{pro:sect2_1}.

\subsection{Proof of Proposition~\ref{pro:seq2_2}}

Recall that in \cite[Proposition 2]{Ha-Ni:2018} it was proved that random matrices ${\bf M}^{(N)}_{L,\sigma^2}(s)=({\bf J}^{(N)}_{L,\sigma^2}(s))^T{\bf J}^{(N)}_{L,\sigma^2}(s)$ and $\widetilde{{\bf M}}^{(N)}_{L,\sigma^2}=(\widetilde{{\bf J}}^{(N)}_{L,\sigma^2})^T\widetilde{{\bf J}}^{(N)}_{L,\sigma^2}$ are equal in distribution for any $s\in\mathbb{R}^N$, where ${\bf J}^{(N)}_{L,\sigma^2}(s)$ and $\widetilde{{\bf J}}^{(N)}_{L,\sigma^2}$ are defined in \eqref{JDW} and \eqref{JDW_tilde}. This implies that random variables $\|{\bf J}^{(N)}_{L,\sigma^2}(s)\|$ and $\|\widetilde{{\bf J}}^{(N)}_{L,\sigma^2}\|$ are equal in distribution, and thus we are left to prove the convergence 
\begin{equation}
\|\widetilde{{\bf J}}_{L,\sigma^2}^{(N)}\|\to \sqrt{\mu_+^{(\infty)}}\quad \text{a.s. as $N\to\infty$}.
\end{equation}

We can use the result of \cite{Pau-Sil:2009} inductively on $L$ to obtain that $(\widetilde{{\bf J}}^{(N)}_{L,\sigma^2})^T\widetilde{{\bf J}}^{(N)}_{L,\sigma^2}$ has no eigenvalues outside of the support of a measure $F^{c_n,A_n,B_n}$ defined in \cite{Pau-Sil:2009}. Then we can use the results of \cite{Cou-Hac:2014} (see Proposition~3.1, Proposition~3.2 and Subsection~3.3) to show that the right edge of $F^{c_n,A_n,B_n}$ converges to $\sqrt{\mu_+^{(\infty)}}$ almost surely as $n\to\infty$. This observation finishes the proof.

\section{Appendix}

\subsection*{A: Proof of Theorem~\ref{thm:main_glob}, LeakyReLU case $0<\alpha<1$}

We consider $L$-layer autoencoder function $X_{N,L}(s)$ with no biases, LeakyReLU activation, and Gaussian initialization of weights:
\begin{equation}
{\bf X}_{L,\sigma^2}^{(N)}(s)=\Phi^{(L)}\circ \ldots\circ \Phi^{(2)}\circ \Phi^{(1)},\quad \Phi^{(k)}(s)=\phi_\alpha(W^{(k)} s),
\end{equation}
where $W^{(k)}$ is a $n_{k+1}\times n_k$ with i.i.d.\  entries distributed as $N(0,\dfrac{\sigma^2}{n_k})$ and $\phi_\alpha$ is the coordinatewise version of LeakyReLU with parameter $\alpha\in(0;1)$:
\begin{equation}
\phi_\alpha(x)=
\begin{cases}
x, \text{\quad if }x\ge 0;\\
\alpha x,\text{\quad if }x<0.
\end{cases}
\end{equation}
Define
\begin{equation}
\|{\bf X}_{L,\sigma^2}^{(N)}\|:=\sup_{\|s\|=1} \|{\bf X}_{L,\sigma^2}^{(N)}(s)\|.
\end{equation}
The proof of Theorem~\ref{thm:main_glob} for the case $0<\alpha<1$ follows the exact same steps as the proof in the case $\alpha=0$ described in Section~\ref{sec:thm2}. Here we specify the main differences.

Define neuron vectors and their norms:
\begin{equation}
s^{(0)}:=s\in\mathbb{R}^N,\quad 
s^{(k)}:=\Phi^{(k)}(s^{(k-1)})=\phi_\alpha(W^{(k)} s^{(k-1)})\in\mathbb{R}^{n_{k+1}},\quad
r^{(k)}:=\|s^{(k)}\|.
\end{equation}
Then 
\begin{equation}
\|{\bf X}_{L,\sigma^2}^{(N)}\|=\sup_{s\in\R^N,\ \|s\|=1}\prod_{k=1}^L\dfrac{r^{(k)}}{r^{(k-1)}}.
\end{equation}
Also define normalized neuron vectors $u_k:=\dfrac{s^{(k)}}{r^{(k)}}\in \Sp{n_{k+1}-1}$. Using positive homogeneity of $\phi_\alpha$ one can show that
\begin{equation}
\dfrac{r^{(k)}}{r^{(k-1)}}=\|\phi_\alpha(W^{(k)}u_{k-1})\|,
\end{equation}
and $u_k$'s satisfy the recurrence relation
\begin{equation}
u_0=s,\qquad
u_k=\dfrac{\phi_\alpha(W^{(k)} u_{k-1})}{\|\phi_\alpha(W^{(k)} u_{k-1})\|}.
\end{equation}
Also positive homogeneity of $\phi_\alpha$ implies
\begin{multline}
(W^{(k)} u_{k-1}, \phi_\alpha(u_k))
=\left(W^{(k)} u_{k-1}, \phi_\alpha\Bigl( \dfrac{\phi_\alpha(W^{(k)} u_{k-1})}{\|\phi_\alpha(W^{(k)} u_{k-1})\|}\Bigr)\right)=\\
=\dfrac{1}{\|\phi_\alpha(W^{(k)} u_{k-1})\|}\bigl(W^{(k)} u_{k-1},\phi_\alpha(\phi_\alpha(W^{(k)} u_{k-1}))\bigr).
\end{multline}
Coordinatewise computation shows that $\Bigl(v,\phi_\alpha(\phi_\alpha(v))\Bigr)=\|\phi_\alpha(v)\|^2$, thus
\begin{equation}
(W^{(k)} u_{k-1}, \phi_\alpha(u_k))=
\|\phi_\alpha(W^{(k)} u_{k-1})\|=\dfrac{r^{(k)}}{r^{(k-1)}},
\end{equation}
which gives a formula similar to~\eqref{eq_XNL_prod_1} for $\|{\bf X}_{L,\sigma^2}^{(N)}\|$ in Leaky ReLU case:
\begin{equation}
\|{\bf X}_{L,\sigma^2}^{(N)}\|=\sup_{s\in\Sp{N-1}} \prod_{k=1}^L \bigl(W^{(k)}u_{k-1},\phi_\alpha(u_k)\bigr).
\end{equation}
Using AM-GM similarly to Subsection~\ref{sbsc:thm2_step1}, we obtain
\begin{equation}
\|{\bf X}_{L,\sigma^2}^{(N)}\|\le 
\Bigg(
\dfrac{1}{L}\sup_{\substack{u_k\in\Sp{n_{k+1}-1},\\k=0,1,\ldots,L}}\quad\sum\limits_{k=1}^L(W^{(k)}u_{k-1},\phi_\alpha(u_k))
\Bigg)^L.
\end{equation}

Denote ${\bf u}:=(u_0,u_1,\ldots,u_L)$ and ${\bf S}:=\Sp{n_1-1}\times\Sp{n_2-1}\times\ldots\times \Sp{n_{L+1}-1}$. We have an upper bound for $\|{\bf X}_{L,\sigma^2}^{(N)}\|$:
\begin{equation}
\|{\bf X}_{L,\sigma^2}^{(N)}\|\le 
\Bigg(
\dfrac{1}{L}\sup_{{\bf u}\in{\bf S}} P_{\bf u}
\Bigg)^L, 
\qquad
P_{\bf u}=\sum\limits_{k=1}^L(W^{(k)}u_{k-1},\phi_\alpha(u_k)),
\end{equation}
where $\{P_{\bf u}\}_{{\bf u}\in{\bf S}}$ is a Gaussian process. This derivation completes \textbf{Step 1}.

In \textbf{Step 2} we need to find majorizing Gaussian process $Q_{\bf u}$ for $P_{\bf u}$.
 Consider independent Gaussian vectors $g_0,g_1,\ldots,g_{L-1},h_1,h_2,\ldots,h_L$ such that each $g_k, h_k\in\R^{n_{k+1}}$ has i.i.d.\ entries distributed as $N(0,1)$. Denote
\begin{equation}
Q_{\bf u}:=\sqrt{\frac{\sigma^2}{n_1}}(g_0,u_0)
    +
    \sum_{k=1}^{L-1}
    \left(
        \sqrt{\frac{\sigma^2}{n_{k+1}}}(g_k,u_k)
        +
        \sqrt{\frac{\sigma^2}{n_k}}(h_k,\phi_\alpha(u_k))
    \right)
    +
    \sqrt{\frac{\sigma^2}{n_L}}(h_L,\phi_\alpha(u_L))
\end{equation}
Computations similar to~\eqref{EP-P'_1} and \eqref{EP-P'_2} show that
\begin{multline}
\E\{(P_{\bf u}-P_{\bf u'})^2\}=\\
=
\sum_{k=1}^L \dfrac{\sigma^2}{n_k} (\|u_{k-1}\|^2\|\phi_\alpha(u_k)\|^2-2(u_{k-1},u_{k-1}')(\phi_\alpha(u_k),\phi_\alpha(u_k'))+\|u_{k-1}'\|^2\|\phi_\alpha(u_k')\|^2)=\\
=\sum_{k=1}^L \dfrac{\sigma^2}{n_k} (\|\phi_\alpha(u_k)\|^2-2(u_{k-1},u_{k-1}')(\phi_\alpha(u_k),\phi_\alpha(u_k'))+\|\phi_\alpha(u_k')\|^2)
\end{multline}
since $\|u_k\|=\|u_k'\|=1$.
Observe that $\|\phi_\alpha(u_k)\|, \|\phi_\alpha(u_k')\|\le 1$ since $\alpha\in (0,1)$, thus $(u_{k-1},u_{k-1}')\le 1$, $(\phi_\alpha(u_{k}),\phi_\alpha(u_k'))\le 1$. 
We can use
\begin{equation}
-xy\le 1-x-y\text{\quad for }x,y\le 1
\end{equation}
to obtain
\begin{equation}
-2(u_{k-1},u_{k-1}')(\phi_\alpha(u_k),\phi_\alpha(u_k'))\le 
\|u_{k-1}\|^2+\|u_{k-1}'\|-2(u_{k-1},u_{k-1}')-2(\phi_\alpha(u_k),\phi_\alpha(u_k')).
\end{equation}
Thus
\begin{equation}\label{Leaky_Var_P}
\E\{(P_{\bf u}-P_{\bf u'})^2\}
\le 
\sum_{k=1}^L \dfrac{\sigma^2}{n_l}\Bigl(\|u_{k-1}-u_{k-1}'\|^2+\|\phi_\alpha(u_k)-\phi_\alpha(u_k')\|^2\Bigr).
\end{equation}
Similarly to~\eqref{VarQ_formula}, using independence of $g_k,h_k$ and the fact that $\E\{(g,v)^2\}=\|v\|^2$ for a Gaussian vector $g$ with i.i.d.\ entries $N(0,1)$, we obtain
\begin{equation}\label{Leaky_Var_Q}
\E\{(Q_{\bf u}-Q_{\bf u'})^2\}
=
\sum_{k=1}^L \dfrac{\sigma^2}{n_l}\Bigl(\|u_{k-1}-u_{k-1}'\|^2+\|\phi_\alpha(u_k)-\phi_\alpha(u_k')\|^2\Bigr).
\end{equation}
Relations \eqref{Leaky_Var_P} and \eqref{Leaky_Var_Q} show that
 Gaussian process $Q_{\bf u}$ majorizes $P_{\bf u}$ in terms of~\eqref{eq:SF-assump}. This observation completes \textbf{Step 2}.

In \textbf{Step 3}
we need  to obtain an upper bound on $\E\{\sup\limits_{{\bf u}\in{\bf S}} Q_{\bf u}\}$ 
From the definition of $Q_{\bf u}$ we have
\begin{multline}
\E\{\sup\limits_{{\bf u}\in{\bf S}} Q_{\bf u}\}
=\sqrt{\frac{\sigma^2}{n_1}}\E\left\{\sup_{u_0\in\Sp{n_1-1}}(g_0,u_0)\right\}+\\
+\sum_{k=1}^{L-1}
\sqrt{\frac{\sigma^2}{n_{k+1}}}\E\left\{\sup_{u_k\in\Sp{n_{k+1}-1}}\left((g_k,u_k)+\sqrt{\dfrac{c_{k+1}}{c_{k}}}(h_k,\phi_\alpha(u_k))\right)\right\}+\\
+ \sqrt{\frac{\sigma^2}{n_L}}\E\left\{\sup_{u_L\in\Sp{n_{L+1}-1}}(h_L,\phi_\alpha(u_L))\right\}.
\end{multline}
For the first summand we have
\begin{equation}
\E\left\{\sup_{u_0\in\Sp{n_{1}-1}} (g_0,u_0)\right\}=\E\{\|g_0\|\}\le \Bigl(\E\{\|g_0\|^2\}\Bigr)^{1/2}=\sqrt{n_1}.
\end{equation} 
For other summands the estimations come from the following fact.

\begin{lemma}\label{lem:leaky_sup}
Let $g,h\in\mathbb{R}^n$ be independent Gaussian vectors with i.i.d.\ entries $N(0,1)$, $\rho\ge 0$, $\mu\ge 0$, $\alpha\in(0,1)$. Then
\begin{equation}
\E\left\{\sup_{u\in\Sp{n-1}}\Bigl(\rho\,(g,u)+\mu(h,\phi_\alpha(u))\Bigr)\right\}\le \sqrt{\rho^2+\mu^2\dfrac{1+\alpha^2}{2}}\cdot\sqrt{n}.
\end{equation}
\end{lemma}
The proof of Lemma~\ref{lem:leaky_sup} is placed in Appendix C.

Using Lemma~\ref{lem:leaky_sup} we obtain
\begin{equation}
\E\{\sup\limits_{{\bf u}\in{\bf S}} Q_{\bf u}\}\le 
\sigma\cdot\left(1+\sum\limits_{k=1}^{L-1}\sqrt{1+\dfrac{c_{k+1}}{c_k}\cdot\dfrac{1+\alpha^2}{2}}+\sqrt{\dfrac{1}{c_L}}\sqrt{\dfrac{1+\alpha^2}{2}}\right).
\end{equation}
Finally, application of Sudakov-Fernique inequality to $P_{\bf u}$ and $Q_{\bf u}$ gives us
\begin{equation}
\dfrac{1}{L}\E\left\{\sup_{{\bf u}\in{\bf S}} P_{\bf u}\right\}\le
\dfrac{\sigma}{\beta_\alpha^{(L)}}. 
\end{equation}
and completes \textbf{Step 3}. 
Finally, \textbf{Step 4} (Gaussian concentration) can be done completely similar to Subsection~\ref{sbsc:thm2_step4}.

\subsection*{B: Singularities of implicit analytic function and radius of convergence}

Recall that ${\bf M}^{(N)}_{L,\sigma^2}$ is $N \times N$ nonnegative matrix, $\nu_{\scriptscriptstyle {\bf M}_{L,\sigma^2}}$ is its limiting NCM, $m_{\scriptscriptstyle {\bf M}_{L,\sigma^2}}(z)$ is the moment generating function:
\begin{equation}
m_{\scriptscriptstyle {\bf M}_{L,\sigma^2}}(z)=\sum_{k=1}^\infty a_kz^k,\quad
a_k=\int_{-\infty}^\infty \lambda^k\,d\nu_{\scriptscriptstyle {\bf M}_{L,\sigma^2}}(\lambda),
\end{equation}
$R$ is the radius of convergence of $m_{\scriptscriptstyle {\bf M}_{L,\sigma^2}}(z)$. Let us explain how to find  $R$ using standard singularity analysis of implicitly given analytic function.

Since $\text{supp}\, \nu_{\scriptscriptstyle {\bf M}_{L,\sigma^2}}\subset [0;+\infty)$, we have $a_k\ge 0$. Vivanti-Pringsheim theorem then yields that $m_{\scriptscriptstyle {\bf M}_{L,\sigma^2}}(z)$ has singularity at $z=R$. Also observe that $m_{\scriptscriptstyle {\bf M}_{L,\sigma^2}}(z)$ is monotone increasing on $[0,R)$ and satisfies equation
\begin{equation}
z=\left(\dfrac{2}{\sigma^2}\right)^L
\dfrac{m_{\scriptscriptstyle{\bf M}_{L,\sigma^2}}(z)}{(m_{\scriptscriptstyle{\bf M}_{L,\sigma^2}}(z)+1)\cdot\prod\limits_{k=1}^L (2c_k m_{\scriptscriptstyle{\bf M}_{L,\sigma^2}}(z)+1)},
\end{equation}
thus there exist $m_*=\lim\limits_{z\to R^-} m_{\scriptscriptstyle {\bf M}_L}(z)$, $m_*\in(0;+\infty)$. Passing to the limit as $z\to R^-$ in~\eqref{m_z_eq_F}, we obtain
\begin{equation}
\mathcal{F}(m_*,R)=0.
\end{equation}
If $\dfrac{\partial \mathcal{F}}{\partial m}(m_*,R)\neq 0$, then implicit function theorem implies that there exist unique analytic solution $m(z)$ of~\eqref{m_z_eq_F} in the neighborhood of $z=R$ satisfying $m(R)=m_*$. Clearly, $m(z)$ is analytic continuation of $m_{\scriptscriptstyle {\bf M}_{L,\sigma^2}}(z)$ in the neighborhood of $z=R$, which is an obvious contradiction to singularity of $z=R$ for $m_{\scriptscriptstyle {\bf M}_{L,\sigma^2}}(z)$. As a conclusion, we proved existence of $m_*>0$ such that $m=m_*$, $z=R$ is a solution of 
\begin{equation}
\begin{cases}
\mathcal{F}(m,z)=0;\\
\dfrac{\partial\mathcal{F}}{\partial m}(m,z)=0.
\end{cases}
\end{equation}

\subsection*{C: Proof of Lemma~\ref{lem:leaky_sup}}

Observe that for $u\in\Sp{n-1}$ we have
\begin{multline}
\rho\,(g,u)+\mu(h,\phi_\alpha(u))=
\sum_{u_k\ge 0} (\rho\, g_k+\mu \,h_k)u_k+\sum_{u_k<0} (-\rho\, g_k-\mu\alpha\, h_k)|u_k|\le \\
\le
\sum_{k=1}^n \max\{\rho\, g_k+\mu\, h_k;-\rho\, g_k-\mu\alpha\, h_k;0\}|u_k|\le 
\left(\sum_{k=1}^n(\max\{\rho\, g_k+\mu\, h_k;-\rho\, g_k-\mu\alpha\, h_k;0\})^2 \right)^{1/2}.
\end{multline}
Denote $g_k':=\rho\, g_k+\mu\, h_k$, $g_k'':=-\rho\, g_k-\mu\alpha\, h_k$, then $g_k'\sim N(0,\rho^2+\mu^2)$, $g_k''\sim N(0,\rho^2+\mu^2\alpha^2)$. Application of Cauchy-Schwarz for expectation provides us
\begin{equation}
\E\left\{\sup_{u\in\Sp{n-1}}\Bigl(\rho\,(g,u)+\mu(h,\phi_\alpha(u))\Bigr)\right\}\le 
\left(\sum_{k=1}^n\E\left\{(\max\{g_k';g_k'';0\})^2\right\} \right)^{1/2}.
\end{equation}
One can check that $(\max\{x,y,0\})^2\le (x_+)^2+(y_+)^2$. Since
\begin{equation}
\E\{((g_k')_+)^2\}=\dfrac{1}{2}\E\{(g_k')^2\}=\dfrac{1}{2}(\rho^2+\mu^2),\qquad
\E\{((g_k'')_+)^2\}=\dfrac{1}{2}(\rho^2+\mu^2\alpha^2),
\end{equation}
we finally obtain
\begin{equation}
\E\left\{\sup_{u\in\Sp{n-1}}\Bigl(\rho\,(g,u)+\mu(h,\phi_\alpha(u))\Bigr)\right\}\le 
\left(n\cdot\Bigl(\dfrac{1}{2}(\rho^2+\mu^2)+\dfrac{1}{2}(\rho^2+\mu^2\alpha^2)\Bigr) \right)^{1/2},
\end{equation}
which completes the proof of Lemma~\ref{lem:leaky_sup}.

\end{document}